\documentclass{article}
\usepackage{arxiv_modern}
\usepackage[utf8]{inputenc} 
\usepackage[T1]{fontenc}    
\usepackage{hyperref}       
\usepackage{url}            
\usepackage{booktabs}       
\usepackage{amsfonts}       
\usepackage{nicefrac}       
\usepackage{microtype}      
\usepackage{xcolor}         
\usepackage{subcaption}
\usepackage{graphicx}
\usepackage{authblk}
\usepackage{siunitx}
\usepackage{placeins}

\title{Adapting Vision-Language Models for Neutrino Event Classification in High-Energy Physics}

\author[1]{Dikshant Sagar}
\author[1]{Kaiwen Yu}
\author[2]{Alejandro Yankelevich}
\author[2,*]{Jianming Bian}
\author[1]{Pierre Baldi}

\affil[1]{Department of Computer Science, University of California, Irvine, CA, USA}
\affil[2]{Department of Physics, University of California, Irvine, CA, USA}
\affil[*]{\textit{Corresponding author:} \url{bianjm@uci.edu}}

\begin{document}

\maketitle

\begingroup
\renewcommand\thefootnote{}
\footnotetext{Accepted for publication in Communications Physics (Nature Portfolio)}
\endgroup

\begin{abstract}
Recent advances in Large Language Models (LLMs)\cite{chang2024survey} have demonstrated their remarkable capacity to process and reason over structured and unstructured data modalities beyond natural language \cite{wu2023multimodal}. In this work, we explore the applications of Vision Language Models (VLMs), specifically a fine-tuned variant of LLaMA 3.2 \cite{grattafiori2024llama} to the task of identifying neutrino interactions in pixelated detector data from high-energy physics (HEP) experiments. We benchmark this model against a state-of-the-art convolutional neural network (CNN) architecture, similar to those used in major neutrino experiments \cite{ayres2007nova, abi2020neutrino, falcone2022deep}, which have achieved high efficiency and purity in classifying electron and muon neutrino events, and also a Vision Transformer (ViT-h/14)\cite{dosovitskiy2020image}, which is the same architecture inside the VLM's vision encoder. Our evaluation considers both classification performance and interpretability of the model predictions, comparing a VLM with a vision-only transformer (ViT) and a convolutional neural network (CNN) baseline. We find that transformer-based architectures outperform conventional CNNs in classification accuracy and robustness, with the VLM providing additional flexibility through the integration of auxiliary textual or semantic information and enabling more interpretable, reasoning-based predictions. These results highlight the potential of large transformer models, particularly vision–language models, as general-purpose backbones for physics event classification, combining strong performance, robustness, and interpretability, and opening new avenues for multimodal reasoning in experimental neutrino physics.


\end{abstract}

\section{Introduction}

Recent years have witnessed a surge in the adoption of machine learning across the physical sciences, driven by unprecedented volumes of experimental data and the promise of uncovering subtle patterns beyond the reach of traditional analyses. In high-energy physics (HEP), this trend is particularly evident: experiments generate vast streams of complex, high-dimensional detector outputs, making automated methods essential for transforming raw observations into scientifically meaningful insights\cite{robles2025particle, yankelevich2024sparse, fenton2024reconstruction,baldi2014searching,baldi2016jet,baldi2016parameterized, chappell2022application}. However, as the field increasingly turns to deep learning techniques, a persistent challenge remains: many of these models, while powerful, operate as opaque black boxes whose predictions are difficult to interpret and validate in a physics context \cite{baldi2021deep}.

A key example of this challenge arises in event classification, where the goal is to distinguish signal interactions of interest from a dominant background. For example, the ability to determine the flavor of neutrinos interacting in a detector is crucial for neutrino oscillation experiments, which aim to measure the rate at which neutrinos of certain flavors convert to different flavors along their trajectory between the source and detector. Historically, this event classification task has relied on first reconstructing higher-level objects within the detector, including resulting particle tracks and showers, and then summarizing their properties through a carefully selected set of engineered features \cite{backhouse2015library}. These features, capturing energies, spatial configurations, and shape descriptors, have served as inputs to algorithms ranging from K-Nearest Neighbors and Boosted Decision Trees to shallow neural networks. While this approach has delivered strong results over decades of experimentation, it also has critical drawbacks: reconstruction errors can degrade classification performance, and the reliance on predefined features constrains the richness of information accessible to the model.

This paradigm echoes the trajectory of computer vision research. For many years, computer vision depended on handcrafted feature extraction pipelines to identify salient characteristics in images. The advent of deep convolutional neural networks (CNNs) fundamentally changed this landscape by enabling models to learn hierarchical representations directly from raw pixel data, outperforming traditional methods and opening new frontiers in visual understanding\cite{abi2020neutrino,lecun1998convolutional,baldi2021deep, chappell2022application, ayres2007nova, falcone2022deep}. Inspired by this progress, researchers in HEP have begun exploring deep learning architectures capable of processing detector data in similarly direct ways \cite{aurisano2016convolutional,baldi2014searching,baldi2016jet,baldi2016parameterized}.

Building on the success of convolutional architectures, recent developments in computer vision have further shifted toward transformer-based models, particularly vision transformers (ViTs)\cite{dosovitskiy2020image}, which replace local convolutional operations with global self-attention mechanisms\cite{vaswani2017attention}. By modeling long-range spatial dependencies directly, ViTs relax the strong locality and translational invariance assumptions inherent to CNNs, enabling more flexible representations of global image structure. This architectural evolution is especially relevant for detector-based imaging tasks in high-energy physics, where physically meaningful features such as extended particle tracks, electromagnetic showers, and interaction vertices often span large spatial regions and exhibit non-local correlations \cite{yankelevich2024sparse}.

In response, transformer-based models have begun to see increasing adoption across a range of HEP applications, including jet tagging, tracking, calorimeter reconstruction, and event-level classification \cite{qu2022particle,baldi2016jet,shmakov2023interpretable}. Their ability to capture sparse, topology-driven patterns and maintain robustness under variations in detector resolution makes ViTs a compelling alternative to purely convolutional approaches. In the context of neutrino event classification, ViTs provide an important intermediate baseline between CNNs and fully multimodal vision-language models, allowing the respective roles of architectural inductive bias and multimodal supervision to be disentangled.

Most recently, Vision Language Models (VLMs), which are large neural networks pretrained on paired visual and textual data, have emerged as a promising extension of these ideas. By jointly learning to associate image content with semantic information, these models can capture nuanced relationships and provide richer, more interpretable embeddings \cite{zhang2024vision}. In the context of neutrino physics, where events can be represented as structured images or tensors and accompanied by labels or descriptions, VLMs offer an exciting opportunity to move beyond conventional pipelines. In addition to improving classification performance, these models could also generate natural-language explanations rooted in knowledge of the underlying physics processes, explicitly referencing event topologies such as muon tracks or electromagnetic showers, which may help elucidate why a particular prediction was made and offer a path toward greater transparency and trust in machine learning-driven analyses.

In this work, we investigate fine-tuning VLMs for event classification in high-energy physics neutrino experiments. Specifically, we consider this task in the context of a liquid argon time projection chamber (LArTPC), a relatively new particle detector technology known for its very high spatial and energy resolution. Our approach leverages the expressive capabilities of VLMs to extract features directly from low-level detector representations, reducing dependence on manually engineered variables. We show that with suitable adaptation, VLMs can deliver strong classification performance and offer new avenues for interpreting complex event signatures in neutrino detectors. In particular, we compare their performance against a conventional CNN \cite{abi2020neutrino} and a ViT \cite{dosovitskiy2020image} and demonstrate that VLMs not only achieve superior classification accuracy but also provide a broader scope of reasoning and more informative explanations for their predictions based on post-hoc autoregressive text generation. Finally, we demonstrate the ability of these VLMs to generalize beyond the specific datasets they are trained on and maintain high performance even under significantly degraded detector conditions, highlighting their robustness and adaptability. These results therefore suggest it would be possible to establish a reusable HEP foundation model, where future adaptations can be achieved even across experiments with minimal further fine-tuning.

\section{Methods}

\subsection{Dataset}
The dataset is a custom simulation of a modular LArTPC with square \SI{5}{mm} pixel-based readout. The detector is $\SI{2}{m} \times \SI{2}{m} \times \SI{7}{m}$ in $x,y,z$ with anodes at $x=\{\SI{-0.9}{m},\SI{-0.3}{m},\SI{0.3}{m},\SI{0.9}{m}\}$ and cathodes at $x=\{\SI{-0.6}{m},\SI{0.0}{m},\SI{0.6}{m}\}$ resulting in \SI{0.3}{m} drift lengths along $x$. Electron neutrino ($\nu_e$) and muon neutrino ($\nu_\mu$) interactions are simulated with GENIE (v3.0.6)~\cite{Andreopoulos:2009rq,Andreopoulos:2015wxa} in the $+z$ direction with uniform neutrino energy up to \SI{10}{GeV}. The dataset consists of 190,000 $\nu_e$ and $\nu_\mu$ events, each with 74\% of events interacting through the charged current and the rest through the neutral current, for which the neutrino flavor cannot be determined and is therefore a significant background for neutrino oscillation experiments. The energy deposition in liquid argon is then simulated with GEANT4 (v11.2.0)  \cite{Geant:2017ats, Agostinelli:2002hh}. To approximate the effect of drift electron transportation in liquid argon \cite{lbl_calc,lbl_diffusion}, the energy deposition in each $\SI{1}{mm} \times \SI{5}{mm} \times \SI{5}{mm}$ voxel is smeared with a Gaussian filter of width \SI{1.3}{mm} (\SI{0.9}{mm}) in the transverse (longitudinal) direction per meter of drift distance to the anode. To generate images for training, two 2D event displays corresponding to XZ and YZ views are made with $\SI{5}{mm} \times \SI{5}{mm}$ pixels. Finally, we crop each event display to a $512 \times 512$ grayscale image (``pixel map'') centered on the interaction, creating the final dataset for training.

\subsection{LLaMA 3.2 Vision}
\begin{figure}[!h]
    \centering
    \includegraphics[width=1.\linewidth]{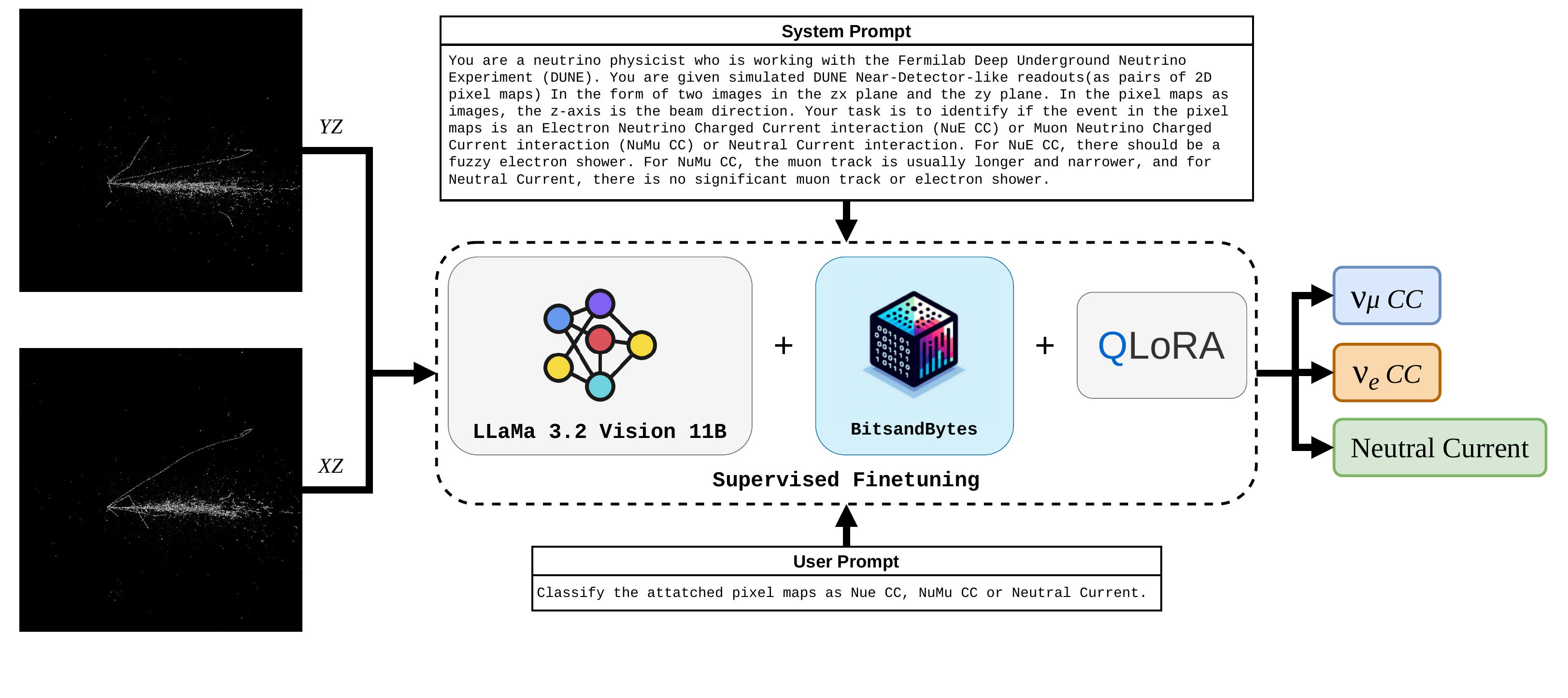}
    \caption{\textbf{LLaMA 3.2 Vision Model Finetuning Overview: } Fine-tuning overview of the LLaMA 3.2 Vision Model for neutrino event classification. Pixel map projections (YZ and XZ) are provided as input, combined with a physics-informed system prompt, and used in a supervised fine-tuning pipeline with BitsAndBytes and QLoRA to classify events into $\nu_\mu$ charged current, $\nu_e$ charged current, or neutral current categories.}
    \label{fig:llamapipeline}
\end{figure}
LLaMA Vision 3.2 is a suite of multimodal large language models developed by Meta, extending the LLaMA 3.2 series with visual capabilities  \cite{grattafiori2024llama}. Unlike traditional CNNs tailored specifically for image-based tasks, LLaMA Vision 3.2 integrates both textual and visual modalities within a unified transformer-based architecture\cite{vaswani2017attention}. It is trained on a diverse corpus of images and documents, enabling it to handle visual inputs such as photographs, rendered plots, and pixelated detector data alongside natural language.
The model utilizes a high-resolution ViT-h/14\cite{dosovitskiy2020image} vision encoder that tokenizes images into patch embeddings, which are then processed alongside text tokens by a shared transformer decoder. This allows for contextual reasoning across modalities, making the model well-suited for tasks that benefit from both visual understanding and symbolic reasoning, such as neutrino event classification in sparse detector images.
In this work, we fine-tune the 11 billion parameter version of LLaMA Vision 3.2 using supervised instruction tuning and a parameter-efficient method known as QLoRA \cite{dettmers2023qlora} on a labeled dataset of neutrino interaction pixel maps. This is visualized as a pipeline in Figure~\ref{fig:llamapipeline}. This allows the model to learn physics-specific features while retaining its pretrained multimodal capabilities. One key advantage of this approach is the model’s ability to produce not just classifications, but also textual justifications or descriptions of events, which can aid interpretability and experimental insight.
By leveraging the flexibility and reasoning capabilities of LLaMA Vision 3.2, we aim to evaluate whether VLMs can serve as competitive or complementary alternatives to conventional CNN and ViT-based approaches in high-energy physics.

\subsubsection{Parameter Efficient Supervised Finetuning}
Fine-tuning large vision-language models like LLaMA Vision 3.2 \cite{grattafiori2024llama} requires significant computational resources due to their billions of parameters and high memory footprint. Fully fine-tuning all model weights is often infeasible, especially when working with domain-specific datasets that are relatively small and do not justify extensive retraining. Moreover, full fine-tuning can lead to overfitting and catastrophic forgetting of pretrained knowledge, particularly in specialized tasks such as neutrino interaction classification using sparse detector images \cite{hu2022lora}.
To address these challenges, as shown in Figure~\ref{fig:llamapipeline}, we adopt a parameter-efficient fine-tuning (PEFT) method, which enables task adaptation by training only a small subset of additional parameters while keeping the majority of the model frozen. Among various PEFT techniques, we employ QLoRA (Quantized Low-Rank Adaptation) \cite{dettmers2023qlora} due to its memory efficiency, scalability, and strong empirical performance in both language and vision-language tasks. QLoRA combines two key ideas: (1) Quantization: The base model weights are stored in 4-bit precision, drastically reducing memory usage without significantly impacting performance. (2) Low-Rank Adaptation (LoRA) \cite{hu2022lora}: Trainable low-rank matrices are injected into the attention and MLP modules, enabling effective task-specific learning with a small number of parameters.
By leveraging QLoRA \cite{dettmers2023qlora}, we are able to fine-tune LLaMA Vision 3.2 11B \cite{grattafiori2024llama} on our neutrino dataset using modest GPU resources while preserving the general visual-linguistic reasoning capabilities of the original model. This approach enables faster iteration, reduced hardware demands, and easier experimentation, making it a practical strategy for applying large models in high-energy physics contexts where computational resources may be constrained.

\subsubsection{Model and Training Specifications}
We fine-tune the LLaMA 3.2 Vision Instruct 11B model, a state-of-the-art multimodal large language model developed by Meta \cite{grattafiori2024llama}. This model combines a high-capacity transformer-based language decoder \cite{vaswani2017attention} with a vision transformer (ViT)-style encoder \cite{dosovitskiy2020image}, enabling joint processing of pixel-level visual data and textual instructions. The Instruct variant is specifically optimized for instruction-following, allowing us to formulate our event classification task as a multimodal prompt-response problem.

We use the \texttt{meta-llama/Llama-3.2-11B-Vision-Instruct} checkpoint, loaded with 4-bit quantization via the \texttt{BitsAndBytes} library to reduce GPU memory usage. The quantization setup is as follows:
\begin{verbatim}
BitsAndBytesConfig(
    load_in_4bit=True, bnb_4bit_use_double_quant=True,
    bnb_4bit_quant_type="nf4", bnb_4bit_compute_dtype=torch.bfloat16
)
\end{verbatim}
This configuration enables us to fine-tune the model on four NVIDIA A6000 GPUs (49GB VRAM each) with a batch size of 4 per device with a balanced distributed strategy.
To make fine-tuning more feasible on large models with limited resources, we employ QLoRA \cite{dettmers2023qlora}, a parameter-efficient method that trains a small number of injected low-rank matrices while keeping the base model frozen. Our QLoRA configuration includes:
\begin{verbatim}
LoraConfig(
    lora_alpha=16, lora_dropout=0.05, r=8, bias="none",
    target_modules=["q_proj", "k_proj", "v_proj", "o_proj",
                    "gate_proj", "up_proj", "down_proj"],
    task_type="VISION_MODEL", 
)
\end{verbatim}

We use Hugging Face's \cite{wolf2019huggingface} \texttt{SFTTrainer} for supervised fine-tuning, with training hyperparameters optimized for stability and efficiency. The model was fine-tuned for a single epoch using a batch size of 4 per device and gradient accumulation of 2, resulting in an effective batch size of 8. We used the \texttt{adamw\_torch\_fused} optimizer with a constant learning rate of $2 \times 10^{-4}$, a warm-up ratio of 0.03, and a maximum gradient norm of 0.3 to ensure stable updates. \texttt{bfloat16} precision with \texttt{TF32} fallback was employed to balance performance and numerical stability. Model checkpoints were saved every 500 steps, and training logs were recorded every 10 steps using TensorBoard. Using the training dataset comprising 190,000 events, the training run was completed in approximately one week.

\subsubsection{Inference}
\begin{figure}[!h]
    \centering
    \includegraphics[width=1.0\linewidth]{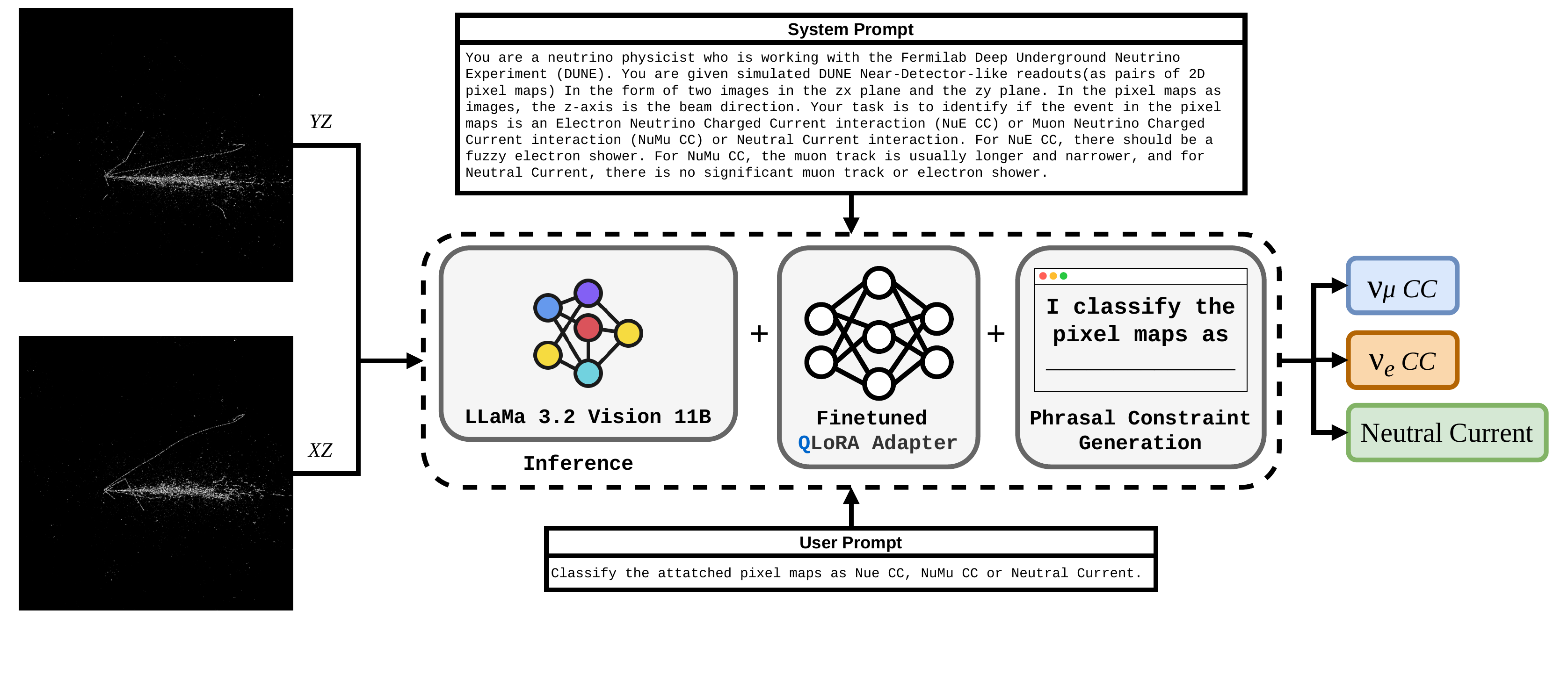}
    \caption{\textbf{LLaMA 3.2 Vision Model Inference Overview: }Inference pipeline for the fine-tuned LLaMA 3.2 Vision Model. YZ and XZ pixel map projections from the detector are processed with a physics-informed system prompt, passed through the base model with a fine-tuned QLoRA adapter, and decoded using constrained generation to produce classifications of $\nu_\mu$ charged current, $\nu_e$ charged current, or neutral current events.}
    \label{fig:llamainferencepipeline}
\end{figure}
For model evaluation, we performed inference on an independent held-out test set comprising 5\% of the dataset samples (10,000 events). Each sample consists of a pair of 2D pixel map images representing orthogonal views in the zx and zy planes. The model was loaded from the base weights (\texttt{meta-llama/Llama-3.2-11B-Vision
-Instruct}) and further initialized with custom adapters we finetuned with our dataset.  During inference, the model received a standardized system message that provided physics-specific context and described the distinguishing features of each interaction class and a user message instructing it to classify each event as one of three categories: electron neutrino charged current ($\nu_{e}$ CC), muon neutrino charged current ($\nu_{\mu}$ CC), or neutral current (NC) interactions similar to the finetuning stage, as shown in Figure~\ref{fig:llamainferencepipeline}.

LLMs or VLMs predict text by autoregressively generating one token at a time, conditioning each new token on the input prompt as well as all previously generated tokens. Given an input prompt and accompanying visual information, the model outputs a probability distribution over the vocabulary for each decoding step, selecting the most likely tokens sequentially \cite{chang2024survey,wu2023multimodal}. However, unconstrained generation can produce variable or verbose phrasing inconsistent with standardized class labels, complicating automated parsing and evaluation. To mitigate this, we applied phrasal constraints, which, under the hood, run a constrained beam search during decoding \cite{hokamp2017lexically}. Specifically, we enforced that the output must begin with a fixed phrase, "I classify the pixel maps as," followed by a token sequence corresponding exclusively to one of the target class labels ($\nu_{e}$ CC, $\nu_{\mu}$ CC, or NC). This was implemented by specifying the constrained prefix as a sequence of token IDs, ensuring that the beam search decoding process could only proceed along paths consistent with the constraint. As a result, during evaluation, the model was compelled to emit predictions in a consistent, machine-readable format while still leveraging its full generative capacity to condition on the visual features and prompt. This approach reduces variability in output text, simplifies downstream confidence scoring, and improves reproducibility of the inference results.

To quantify model confidence in each prediction, we computed a joint probability distribution over the three target classes. Specifically, after generating the output text, we extracted the logarithmic softmax normalized probabilities corresponding to the first token of each class label at the decoding position immediately following the fixed prompt prefix (“I classify the pixel maps as”). This decoding index is where the model begins emitting the class label itself. For each of the three classes ($\nu_{e}$ CC, $\nu_{\mu}$ CC, and NC), we retrieved the log-probabilities of their respective canonical start tokens at this position. These values reflect the model’s relative preference for each class when it commits to generating the label.

To convert these log-probabilities into normalized class probabilities, we applied a temperature-scaled softmax transformation. Concretely, the vector of log-probabilities was scaled by a scalar $T$ to sharpen the distribution before applying the softmax function across the three classes:
\begin{equation}
    P(C_i) = softmax(T \cdot log \space p(C_i))
\end{equation}
where $P(C_i)$ denotes the final confidence assigned to class $C_i$ and $T=5$.
This procedure yields an interpretable probability distribution over the three classes for each prediction, emphasizing the most likely class while retaining information about the relative likelihoods of the alternatives \cite{petroni2019language, guo2017calibration}.

\subsubsection{Prediction Explainability}
\begin{figure}[!h]
    \centering
    \includegraphics[width=1.0\linewidth]{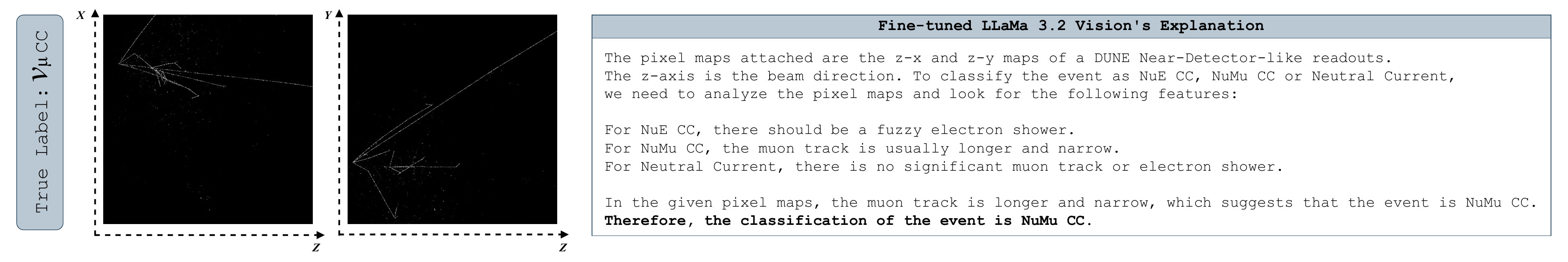}
    \caption{\textbf{Finetuned LLaMa 3.2 Vision's Explanation Example: }Explanation generated by the finetuned LLaMa 3.2 Vision for pixel maps in the x–z and y–z projections for a simulated LArTPC event, labeled as a $\nu_\mu$ charged-current (CC) interaction.}
    \label{fig:llama-explaination}
\end{figure}
In neutrino physics, particularly in the classification of neutrino interaction events from detector pixel maps, interpretability is critical for validating model predictions against established physical understanding. A notable advantage of VLMs over conventional CNNs or ViTs lies in their ability to provide human-readable explanations for their predictions. While CNNs and ViTs primarily output numerical class probabilities or embeddings, their internal decision-making process is opaque, typically requiring post-hoc interpretability tools such as Grad-CAM, saliency maps, feature maps and attention map visualization to approximate the reasoning behind a prediction. These methods can highlight regions of interest in the input image but do not inherently articulate why those regions influence the output \cite{baldi2021deep, selvaraju2017grad,sundararajan2017axiomatic}.

In contrast, VLMs by virtue of their joint vision-language training, can generate natural language rationales that connect visual evidence to semantic concepts. Given an input image and a query, a VLM can not only identify the relevant object or scene but also explain its decision in textual form, often referencing specific visual cues \cite{sammani2022nlx}. For example, a VLM might classify an event as a “muon neutrino charged-current interaction” and generate a textual explanation for detector pixel maps as shown in Figure~\ref{fig:llama-explaination}.  

Consequently, the explanation is grounded in both the visual patterns of the pixel maps and the physics concepts relevant to event topology. While these explanations are not a perfect reflection of the model’s internal causal reasoning, they provide a more accessible and physics-aware interpretability interface than purely visual attribution maps from CNNs. This makes VLMs a promising direction for explainable AI in neutrino event classification.

\subsection{Vision Transformer (ViT)}

Vision Transformers (ViTs) have emerged as a powerful alternative to convolutional architectures for image classification tasks, demonstrating strong performance across a wide range of computer vision benchmarks by modeling long-range dependencies through self-attention mechanisms\cite{vaswani2017attention, dosovitskiy2020image, steiner2021augreg}. Unlike CNNs, which rely on local receptive fields and hierarchical feature extraction, ViTs operate on sequences of image patches and enable global contextual reasoning across the entire input. This characteristic makes ViTs particularly attractive for detector image analysis, where correlations across distant spatial regions can be physically meaningful, such as extended tracks or spatially separated energy deposits\cite{shmakov2023interpretable}.

In high-energy physics applications, ViT-based models have recently been explored for tasks including jet tagging, calorimeter image classification, and neutrino event identification, showing competitive or superior performance relative to CNNs while maintaining architectural simplicity\cite{qu2022particle, shmakov2023interpretable}. In this work, we include a ViT model as a strong vision-only baseline to assess how far purely visual architectures can be pushed before incorporating multimodal reasoning, and to provide a direct comparison with both the CNN baseline and the proposed vision-language model.

\subsubsection{ViT-H/14 Architecture}

\begin{figure}
    \centering
    \includegraphics[width=1\linewidth]{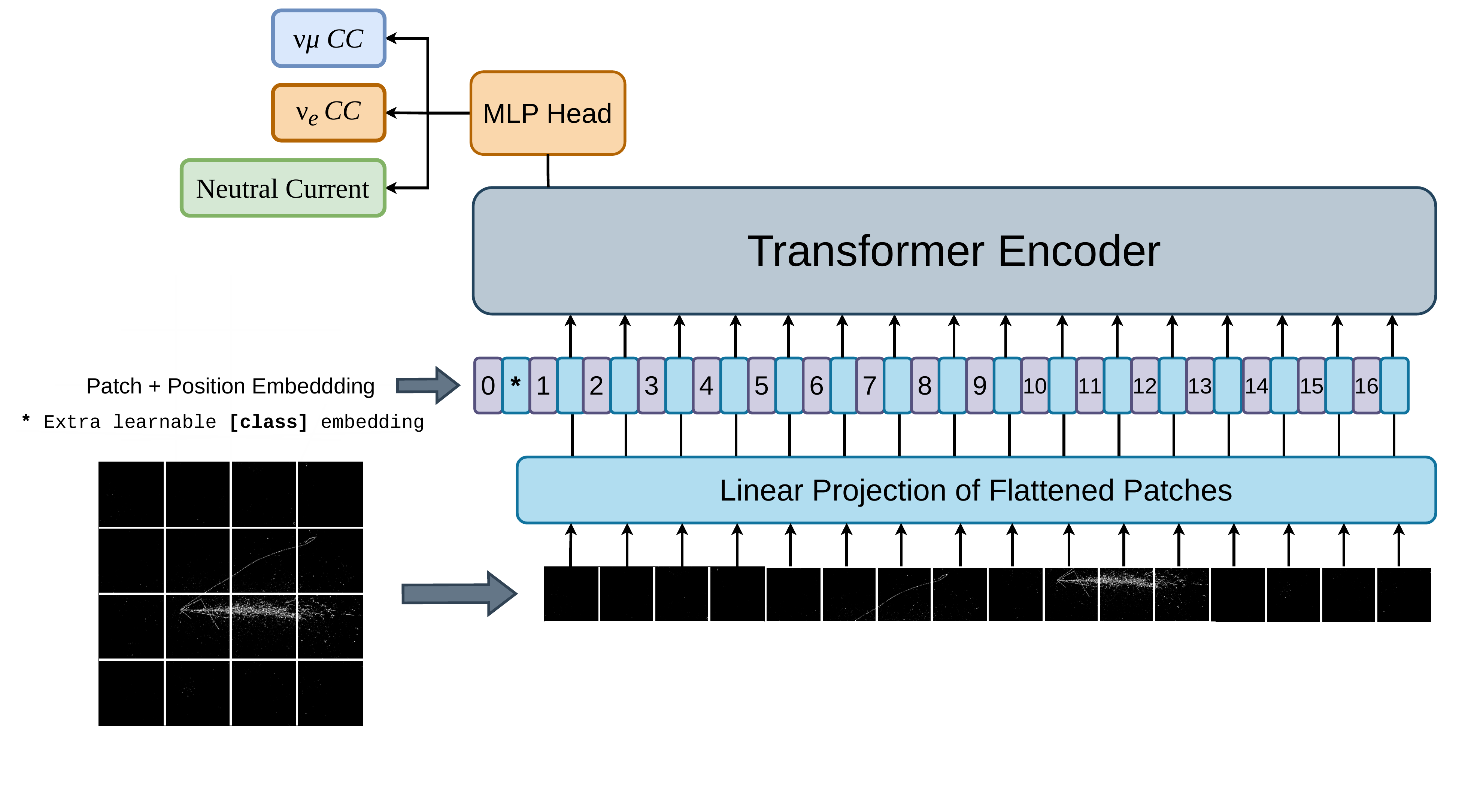}
    \caption{\textbf{ViT-h/14 Architecture: }ViT-h/14 splits an image into 14x14 patches, linearly embeds them, adds positional embeddings, and feeds the resulting sequence of vectors to a standard Transformer encoder, which in turn feeds into a classification MLP head.}
    \label{fig:vith14}
\end{figure}

We adopt the ViT-h/14 architecture\cite{dosovitskiy2020image} as the transformer-based baseline in this study, which corresponds to the vision backbone used within our chosen vision–language model (VLM). The model divides each input image into non-overlapping $14\times14$ patches, which are linearly projected into a sequence of patch embeddings and augmented with learnable positional encodings. These embeddings are processed by a deep stack of transformer encoder blocks, each consisting of multi-head self-attention layers followed by feed-forward networks, layer normalization, and residual connections (See Figure \ref{fig:vith14}).

The ViT-h/14 model contains approximately 632 million trainable parameters when fully fine-tuned. A classification token is prepended to the patch sequence, and its final hidden representation is passed to a linear classification head producing a three-class softmax output corresponding to the neutrino interaction categories used in this analysis. Compared to CNN-based approaches, this architecture enables direct global context aggregation at every layer, without inductive biases toward locality or translation invariance.

\subsubsection{Training Setup}

The ViT-h/14 baseline is trained using full fine-tuning in a supervised learning setting. The input consists of 2 grayscale detector images with a resolution of $512\times512$, which are patchified according to the ViT-h/14 input specification. The model is optimized using the Adam optimizer\cite{kingma2014adam} with an initial learning rate of $1\times10^{-4}$ and a weight decay of 0.05. A per-device batch size of 4 is used during training with gradient accumulation of 4 steps.

The model is trained for 10 epochs; however, the loss saturated pretty quickly due to a large number of training parameters. All parameters of the network are updated during training. Training is performed on 8 NVIDIA A5000 GPUs using PyTorch.

\subsection{CNN}
Convolutional neural networks (CNNs) have long been widely used in problems such as event classification, feature extraction, and image segmentation in high-energy physics image analysis tasks\cite{refId0,aurisano2016convolutional, robles2025particle}. Due to their advantages in local feature modeling and spatial invariance, CNN architectures exhibit good performance in processing sparse pixel maps, detector images, and other visual data\cite{baldi2014searching, acciarri2017convolutional}. However, the expressive power of CNN models is often limited to the visual domain itself and lacks the ability to interpret information at the physical-semantic or symbolic level, which can be a limitation in scientific tasks that require incorporating contextual understanding or providing interpretable output\cite{carleo2019machine}. Therefore, we use CNN as a comparative benchmark in this work to explore its performance on neutrino image data and systematically compare it with our proposed multimodal macromodel, LLaMA Vision 3.2, to assess the latter's potential and advantages in combining visual understanding with textual inference. 

\subsubsection{CNN} Architecture
We developed a multi-branch, SE‑ResNet–style architecture to work as the classification model used in this work. This model uses ResNet‑style residual units\cite{he2016deep} with standard ReLU nonlinearities\cite{agarap2018deep} and explicit SE attention\cite{hu2018squeeze}. Also, it adopts a Siamese architecture\cite{bromley1993signature, koch2015siamese}, where a pair of input images are processed independently through identical sub-networks and later merged for joint reasoning. This CNN baseline contains approximately 21.7 million parameters and is designed to efficiently handle high-resolution, sparsely populated pixel maps. It serves as a representative conventional CNN approach for neutrino event classification, enabling a direct comparison with the vision-language capabilities of LLaMA Vision 3.2.

\begin{figure}[!h]
    \centering
    \includegraphics[width=\linewidth]{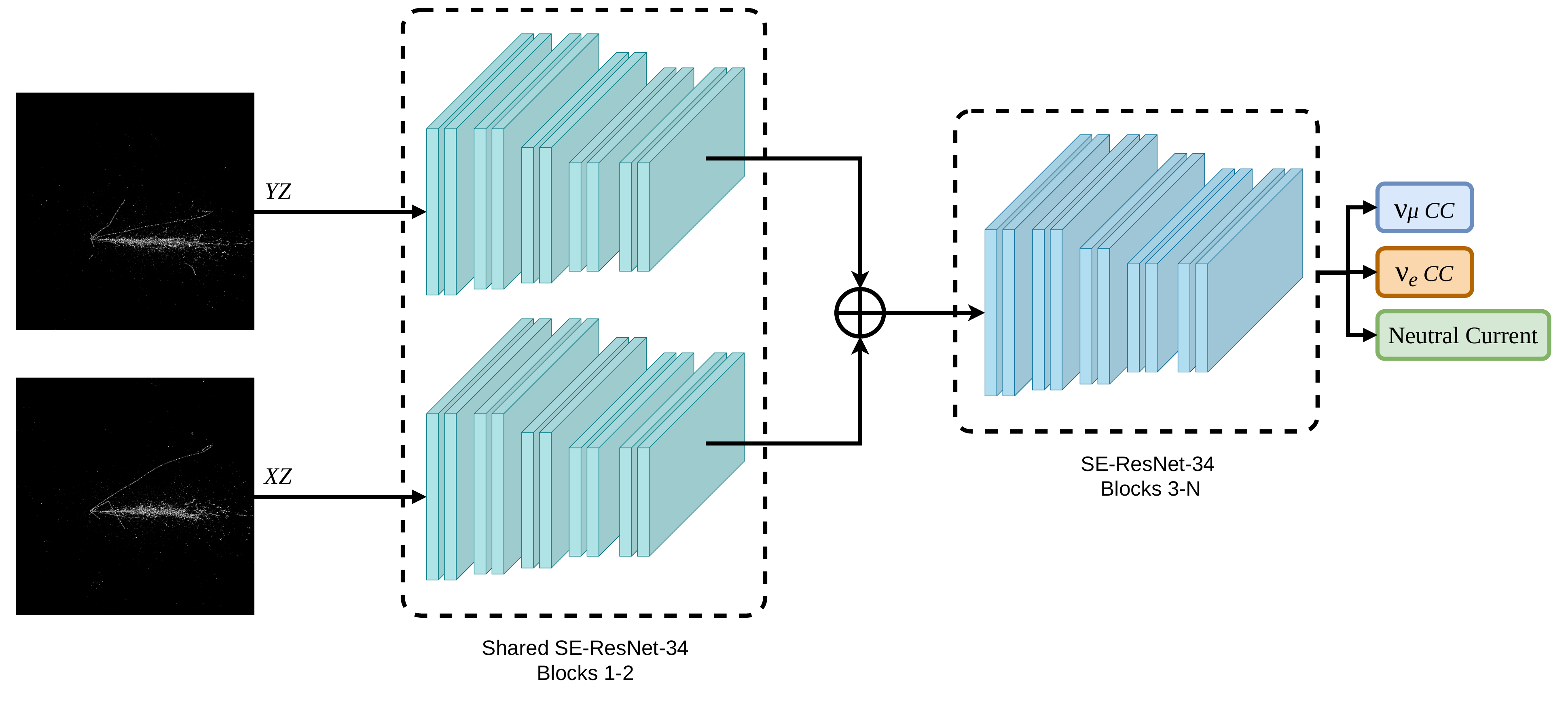}
    \caption{\textbf{CNN Architecture: }Simplified diagram of the CNN architecture based on \cite{abi2020neutrino}. The model takes in pixel maps in the x–z and y–z projections for a simulated LArTPC event and produces an event class output.}
    \label{fig:CNN pipeline}
\end{figure}

The CNN is operated on three 500×500 single‑channel image views per event. Each view is processed by a dedicated branch that begins with an initial 7×7 convolution (stride 2) followed by a sequence of pre‑activation residual blocks (BatchNorm → ReLU → Conv). Squeeze‑and‑Excitation (SE) modules are integrated into several residual blocks to provide channel‑wise feature recalibration. After per‑view feature extraction, branch outputs are concatenated along the channel dimension and passed through additional residual blocks. The merged feature map is globally average pooled and fed to fully connected output heads for the final predictions. The structure of the design is given in Figure~\ref{fig:CNN pipeline}

To make the CNN adapt to the need of our experiment, we changed the input head to take 2 grayscale image with $512\times 512$ resolution. Also, we modified the interaction output head to produce a single 3‑way softmax (three neurons) corresponding to the interaction categories used in our analysis. 



\subsubsection{Training Setup}
We train the convolutional baseline model using a supervised learning framework. The input to the network consists of 2 grayscale detector images with a resolution of $512\times 512$. Each training sample consists of a pair of images corresponding to a neutrino interaction event, processed through a Siamese architecture as described earlier.

The model is optimized using the Adam optimizer\cite{kingma2014adam} with an initial learning rate of $1\times 10^{-6}$ and weight decay of $1\times 10^{-4}$. A batch size of 16 is used during training. The model is trained for up to 300 epochs, with early stopping applied if the validation loss does not improve for 10 consecutive epochs. The objective function is the cross-entropy loss, and all training is performed on a single NVIDIA A6000 GPU using TensorFlow. During the training, the program takes around 26 GB of memory, and takes around 210 minutes for one epoch of training.

\section{Results}
\subsection{Event Classification}
The LLaMA 3.2 Vision-Instruct model demonstrated strong performance in the classification of neutrino interaction events from simulated detector pixel maps. Compared to the CNN baseline, our fine-tuned LLaMA 3.2 Vision consistently achieved higher accuracy, precision, recall, and AUC-ROC and more reliable confidence estimates. Also, relative to the ViT-h/14 baseline, the VLM attained higher accuracy and precision with improved computational efficiency, while achieving comparable AUC-ROC performance.


Table~\ref{metrics_table} summarizes the classification performance and computational characteristics of the evaluated models. While LLaMA~3.2 Vision achieves the highest accuracy, precision, and recall (0.87 each) with an AUC-ROC of 0.96, its performance is comparable to the ViT-h/14 baseline, which attains similar accuracy (0.86), precision (0.86), recall (0.85), and an identical AUC-ROC of 0.96. Importantly, these results are obtained under markedly different training regimes: LLaMA~3.2 Vision is fine-tuned using parameter-efficient training, updating only 29.5M parameters via QLoRA for a single epoch, whereas ViT-h/14 is fully fine-tuned with 632M trainable parameters over 10 epochs. The CNN baseline, trained end-to-end with 21.7M parameters for 300 epochs, exhibits substantially lower classification performance. These results highlight that competitive classification accuracy and discriminative capability can be achieved with substantially fewer trainable parameters and limited fine-tuning when leveraging large pre-trained vision–language models.

\begin{figure}[!h]
    \centering
    \includegraphics[width=1.0\linewidth]{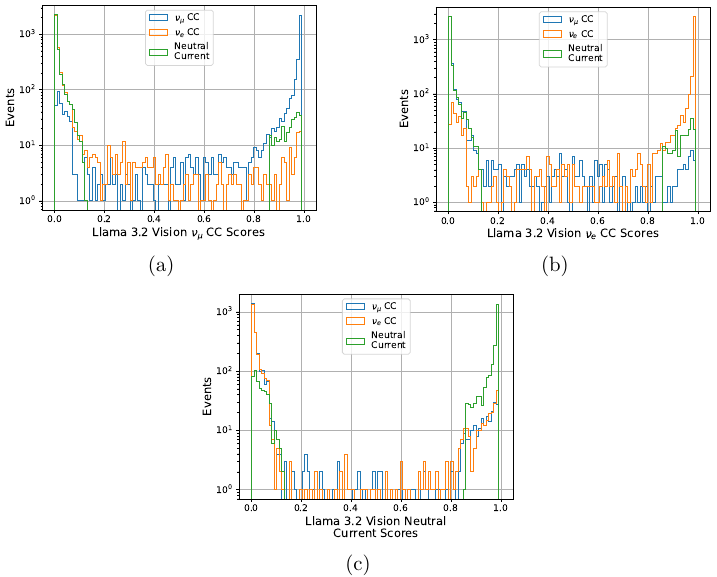}
    \caption{\textbf{LLaMA 3.2 Vision Model Event Classification Rejection Curves: }LLaMA 3.2 Vision model's (a) $\nu_{\mu}$ CC event signal-background rejection curves, (b) $\nu_e$ CC confidence plot, and (c) NC confidence plot. Blue curves belong to $\nu_\mu$ CC events, Orange curves belong to $\nu_e$ CC events, and Green curves belong to Neutral Current events.}
    \label{fig:llama-rejection curve}
\end{figure}

Because LLaMA~3.2 Vision is a generative vision–language model rather than a conventional discriminative classifier, the confidence distributions shown in Fig.\ref{fig:llama-rejection curve} are derived using a non-standard probability estimation procedure explained in the Methods section. As a result, the resulting confidence scores reflect the model’s relative preference for generating a given class label under the imposed decoding constraints, rather than calibrated posterior probabilities in the traditional sense. These distributions should therefore be interpreted qualitatively, as indicators of separability and relative confidence, rather than as direct probabilistic outputs comparable to those of standard neural network classifiers.

An interesting observation we saw was the strong bimodal distribution observed in the NC confidence scores, with events clustering near probabilities of 0 or 1, likely arises from a combination of physical, architectural, and representational factors. Here we list some possible explanations: Physically, NC interactions lack a final-state charged lepton, producing topologies dominated by hadronic activity without the extended tracks or electromagnetic showers characteristic of CC events; when this absence is visually clear, the model assigns high NC confidence, while even partial track- or shower-like features rapidly suppress it. Architecturally, the transformer-based vision encoder aggregates global spatial information, reinforcing a near-binary separation between events with and without salient charged-lepton signatures. In addition, semantic priors associated with the NC class tokens (e.g., “neutral” and “current”) in the language component may interact with learned visual features during fine-tuning, further amplifying confidence polarization. Although the relative contributions of visual evidence and linguistic priors cannot be disentangled here, the resulting scores are consistent with robust discrimination based on high-level event topology and should be interpreted as relative model preferences rather than calibrated probabilities.

\begin{figure}[!h]
    \centering
    \includegraphics[]{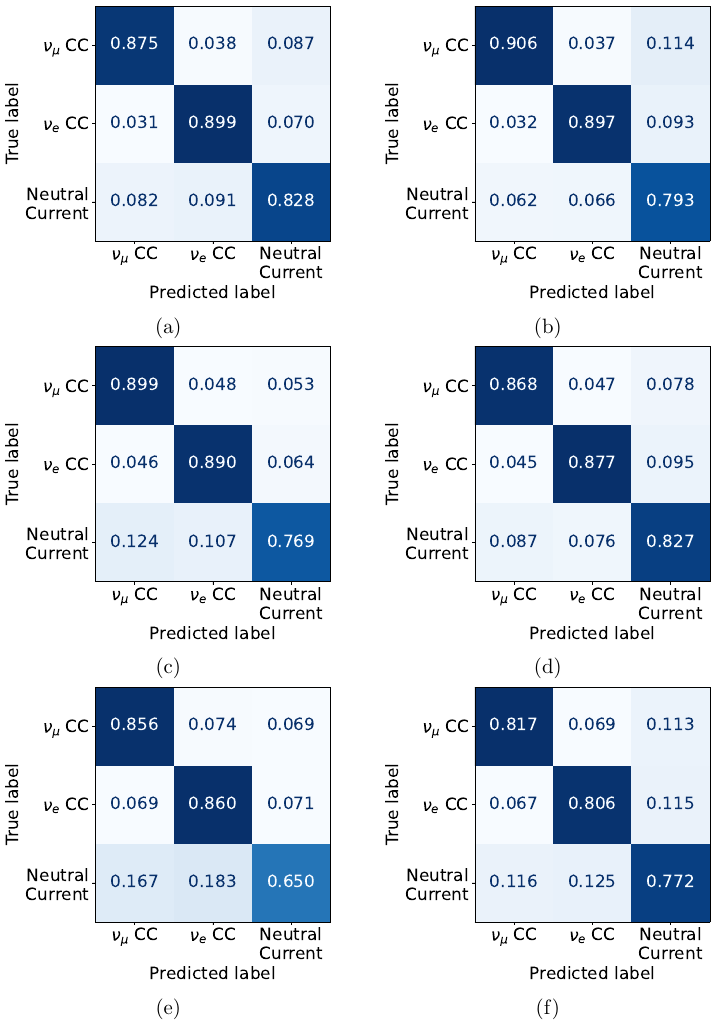}
    \caption{\textbf{Efficiency and Purity Matrices: }Finetuned LLaMA 3.2 Vision's (a) recall matrix (truth normalized) and (b) precision matrix (prediction normalized). ViT-h/14's (c) recall matrix (truth normalized) and (d) precision matrix (prediction normalized).CNN's (e) recall matrix (truth normalized) and (f) precision matrix (prediction normalized).}
    \label{fig:llama-cm}
\end{figure}

\begin{figure}[!h]
    \centering
    \includegraphics[]{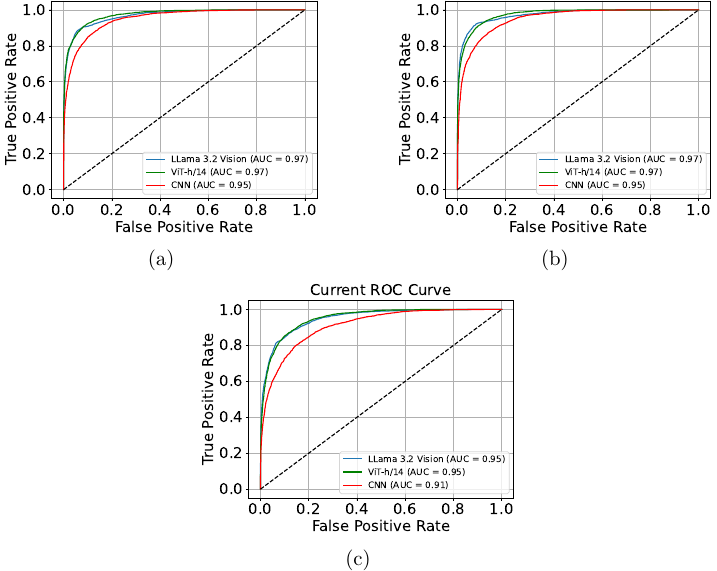}
    \caption{\textbf{AUC-ROC Curves: }ROC curves for each class (a) $\nu_\mu$ CC, (b) $\nu_e$ CC, and (c) NC comparing performance between the finetuned LLaMA 3.2 Vision and the CNN. Blue curves belong to LLama 3.2 Vision (AUCs=0.97,0.97,0.95), green curves belong to ViT-h/14 (AUCs=0.97,0.97,0.95), red curves belong to CNN (AUCs=0.95,0.95,0.91), and the black dotted lines represent a classifier with no predictive power.}
    \label{fig:roc}
\end{figure}

Figure~\ref{fig:llama-cm} shows the confusion matrices of the LLaMA~3.2 Vision, ViT-h/14, and CNN models, respectively, enabling a comparative evaluation of their class-level classification behavior. LLaMA~3.2 Vision demonstrates consistently strong classification performance across all classes, with particularly improved NC identification and enhanced $\nu_{e}$–NC discrimination, which is especially important for neutrino oscillation analyses. While the ViT-h/14 model exhibits comparable overall performance, LLaMA~3.2 Vision shows a more balanced trade-off between efficiency (recall) and purity (precision) across the three interaction classes, resulting in more stable and reliable classification behavior. In contrast, the CNN baseline exhibits increased class confusion, particularly between $\nu_{e}$ and NC events. In terms of discriminative capability, both transformer-based models outperform the CNN baseline. Both LLaMA 3.2 Vision and the ViT-h/14 achieve an AUC-ROC of 0.97, compared to 0.72 for the CNN (See Figure \ref{fig:roc}).

\begin{figure}[!h]
    \centering
    \includegraphics[]{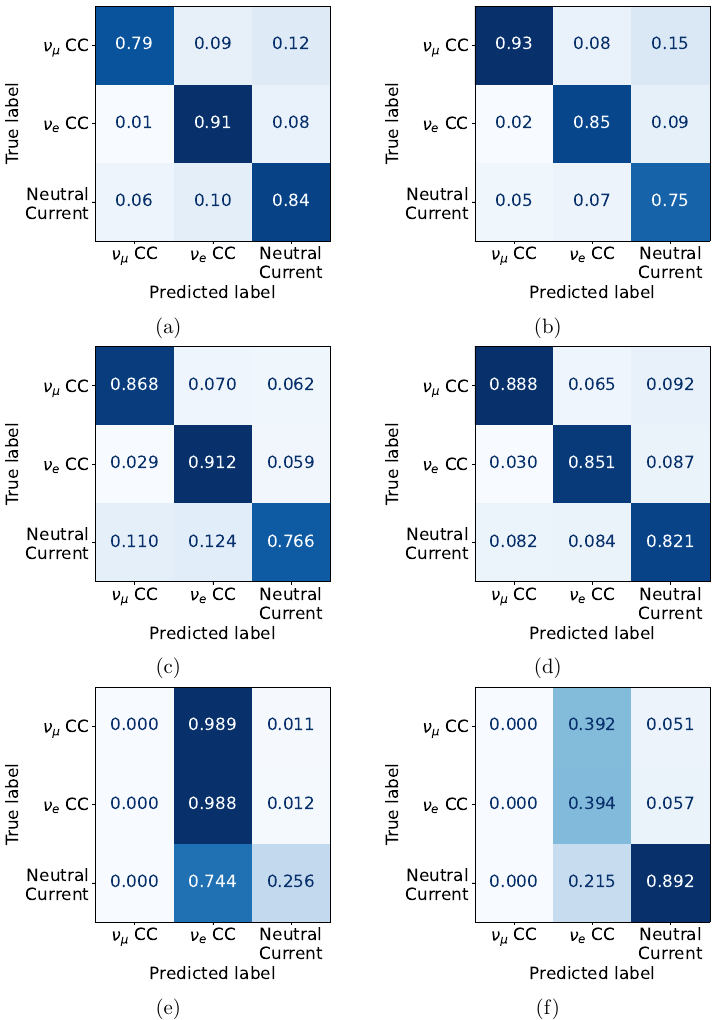}
    \caption{\textbf{Generalization Testing Efficiency and Purity Matrices: }Finetuned LLaMA 3.2 Vision's (a) recall matrix (truth normalized) and (b) precision matrix (prediction normalized) for generalization testing. ViT-h/14's (c) recall matrix (truth normalized) and (d) precision matrix (prediction normalized) for generalization testing. CNN's (e) recall matrix (truth normalized) and (f) precision matrix (prediction normalized) for generalization testing.}
    \label{fig:llama-cm-generalization}
\end{figure}

\begin{figure}[!h]
    \centering
    \includegraphics[]{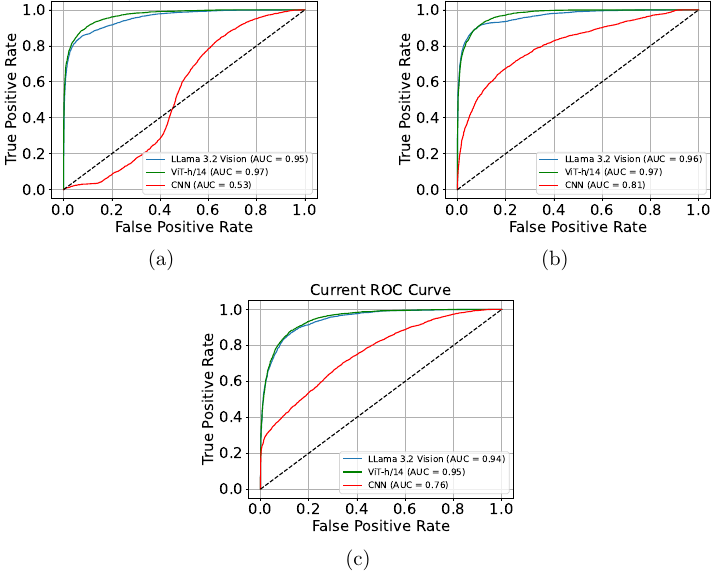}
    \caption{\textbf{Generalization Testing AUC-ROC Curves: }ROC curves for each class (a) $\nu_\mu$ CC, (b) $\nu_e$ CC, and (c) NC comparing performance between the finetuned LLaMA 3.2 Vision and the CNN for generalization testing. Blue curves belong to LLama 3.2 Vision (AUCs=0.95,0.96,0.94), green curves belong to ViT-h/14 (AUCs=0.97,0.97,0.95), red curves belong to CNN (AUCs=0.53,0.81,0.76), and the black dotted lines represent a classifier with no predictive power.}
    \label{fig:generalization-roc}
\end{figure}

\subsection{Generalization Testing}

We also conducted generalization testing by running inference with all models on neutrino event pixel maps downsampled to half the original resolution ($256\times256$). This setting evaluates each model’s ability to maintain performance under reduced spatial detail, mimicking scenarios with lower detector resolution or aggressive data compression. As shown in Table~\ref{metrics_table_generalization}, both LLaMA~3.2 Vision and the ViT-h/14 baseline maintain strong and nearly identical classification performance under this distribution shift, each achieving an accuracy, precision, and recall of 0.85. In contrast, the CNN baseline exhibits a substantial degradation in performance, with accuracy, precision, and recall dropping to 0.49.

In terms of discriminative capability, the transformer-based models continue to outperform the CNN baseline. LLaMA 3.2 Vision achieves an AUC-ROC of 0.95, while the ViT-h/14 model attains a slightly higher AUC-ROC of 0.96, compared to 0.72 for the CNN. We further present the confusion matrices (Figure ~\ref{fig:llama-cm-generalization} and the ROC curves (Figure~\ref{fig:generalization-roc}) for this analysis.

Overall, these results indicate that transformer-based architectures, including both the vision–language model and the vision-only ViT baseline, exhibit greater robustness to spatial downsampling than the CNN. For LLaMA 3.2 Vision, this robustness is achieved while finetuning a substantially less number of parameters and retaining the additional capability to generate post hoc, human-readable explanations, which may be beneficial for downstream analysis and diagnostics in realistic detector conditions.

\subsection{Few-Shot In-Context Evaluation}

To assess whether LLaMA~3.2 Vision can perform neutrino interaction classification without task-specific fine-tuning, we conducted a few-shot in-context evaluation using the frozen pre-trained model. In this setting, the model parameters were kept fixed, and task adaptation was attempted solely through prompt design. Specifically, we provided a single labeled example for each of the three interaction categories ($\nu_e$ CC, $\nu_\mu$ CC, and NC) within the prompt, followed by an unlabeled query event for classification.

Despite this in-context supervision, the model consistently predicted all query events as $\nu_e$ CC interactions. As a result, the few-shot evaluation achieved an overall classification accuracy of 0.3678, corresponding to the class prior of the dominant predicted category. No meaningful class separation was observed across the evaluated events.

These results indicate that, in the absence of fine-tuning, LLaMA~3.2 Vision is unable to reliably map low-level detector pixel representations to the abstract physical interaction categories required for neutrino event classification. This behavior suggests that the visual features learned during pre-training are insufficiently aligned with the domain-specific semantics of sparse detector images, and that parameter adaptation is necessary to bridge this gap. The observed failure mode further motivates the use of parameter-efficient fine-tuning strategies, such as QLoRA, to effectively specialize large vision–language models for scientific imaging tasks with limited labeled data.

\subsection{Event Explainability}
While the LLaMA~3.2 Vision model incurs substantially higher computational overhead, averaging 12.9~GB of inference memory (over 5× the 2.44~GB required by the CNN) and approximately 3.4~seconds of inference time per sample, compared to 23.9~milliseconds for the CNN, its advantages extend beyond raw accuracy alone. The ViT-h/14 baseline occupies an intermediate point in this trade-off space, requiring 2.6~GB of memory and 299~milliseconds per sample while achieving comparable classification performance. In addition to achieving strong classification accuracy on neutrino event pixel maps, LLaMA~3.2 Vision offers an interpretability advantage that CNNs and ViT models inherently lack. Leveraging its vision–language alignment, LLaMA~3.2 Vision can accompany its predictions with natural language explanations grounded in event topology, such as identifying long muon tracks, electromagnetic showers, or the absence of hadronic activity to justify its classification (Figure~\ref{fig:llama-explaination}). This capability allows physicists to assess whether the model’s reasoning is consistent with established physics heuristics, aiding both trust and error diagnosis. Furthermore, although the training and inference costs are higher, these resources enable the development of a reusable foundation model that can be adapted to other detector tasks via lightweight fine-tuning, substantially reducing the effort required for future applications.

These results highlight a trade-off not merely between accuracy and efficiency, but between computational cost and the depth of insight and adaptability provided by different model classes. The CNN remains well-suited for real-time or resource-constrained deployments due to its minimal memory footprint and low latency. The ViT-h/14 model occupies an intermediate position, offering competitive classification performance with moderate computational requirements, making it attractive for large-scale offline processing where throughput remains a consideration. In contrast, LLaMA~3.2 Vision’s richer output, combining strong predictive performance with human-readable, physics-grounded justifications, makes it a compelling choice for offline analyses, detailed event studies, and applications where interpretability and adaptability are as critical as raw accuracy.

\subsection{Ablation Study: Role of Physics Definitions in the System Prompt}

\begin{figure}
    \centering
    \includegraphics[width=1.0\linewidth]{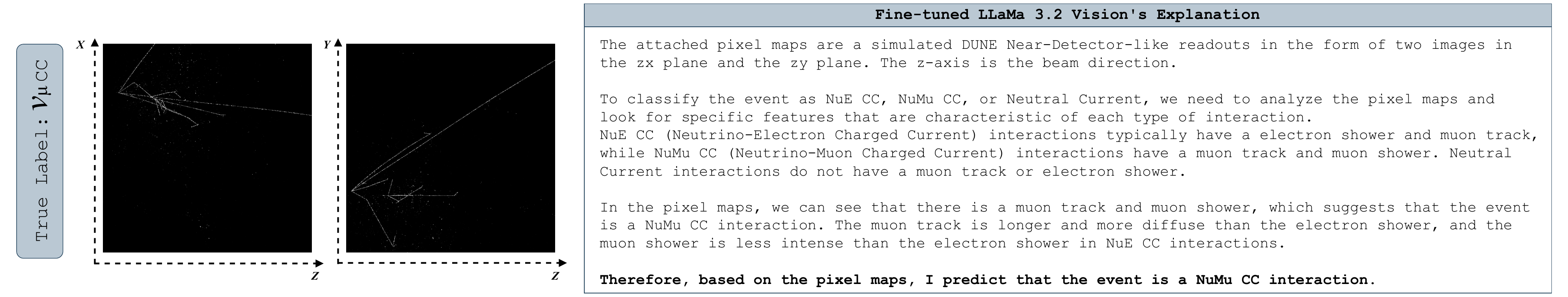}
    \caption{\textbf{Finetuned LLaMa 3.2 Vision’s Explanation Example in the Ablation Setting: }Explanation generated by the finetuned LLaMa 3.2 Vision in the ablation setting for pixel maps in the x–z and y–z projections for a simulated LArTPC event, labeled as a $\nu_\mu$ charged-current (CC) interaction.}
    \label{fig:llama-exp-ab}
\end{figure}

To assess the reliance of the fine-tuned LLaMA~3.2 Vision model on explicit physics guidance, we conducted an ablation study in which the physics definitions and event-type descriptions were removed from the system prompt at inference time. Under this setting, the model was instructed only to perform event classification and to generate an accompanying explanation, without being provided with domain-specific descriptions of neutrino interaction categories.

Qualitatively, the model continues to produce explanations that reference salient visual features of the detector images, such as long track-like structures, localized electromagnetic activity, or the absence of visible charged-lepton signatures. These explanations remain largely consistent with those generated when physics definitions are included in the prompt. Fig. \ref{fig:llama-exp-ab} shows the explanation generated in the ablation setting for the same sample in Fig. \ref{fig:llama-explaination}.

Importantly, we emphasize that these explanations are post hoc in nature and are not claimed to reflect the model’s internal decision-making process. Moreover, they have not yet been evaluated by human experts for physical correctness or usefulness.

Quantitatively, the model achieves an accuracy of 0.86, precision of 0.86, recall of 0.86, and an AUC-ROC of 0.96 under the ablated prompt condition. These results indicate that removing explicit physics definitions from the system prompt does not substantially degrade classification performance, suggesting that the fine-tuned model is able to perform the task using representations learned during pre-training and fine-tuning rather than relying solely on prompt-level domain descriptions.

The persistence of similar post hoc explanations and strong classification performance despite the removal of explicit physics definitions suggests that the VLM can draw upon latent textual knowledge acquired during pre-training to generate plausible physics-related descriptions. This behavior should be interpreted as evidence of the model’s language-generation capabilities rather than as an indication of genuine physical understanding. The generated explanations are therefore best viewed as auxiliary interpretive signals that may aid qualitative inspection and error analysis, rather than as validated explanations of the model’s reasoning. Future work will include a human-in-the-loop evaluation of explanation fidelity, physical correctness, and practical usefulness for domain experts.


\FloatBarrier
\section{Conclusion}

We compared the CNN baseline and ViT-h/14 with the LLaMA~3.2 Vision model for neutrino event classification and observed a clear trade-off between computational efficiency, predictive capability, and explainability. While LLaMA~3.2 Vision demands substantially higher computational resources, averaging 12.9~GB of memory usage and significantly longer inference times compared to the CNN and the ViT-h/14, it consistently delivers superior accuracy across multiple evaluation settings, including challenging generalization scenarios with reduced spatial resolution. We observed that a vision-only transformer (ViT-H/14) achieves classification performance comparable to the vision–language model under both nominal and degraded detector conditions, highlighting the robustness of transformer-based architectures relative to convolutional models.

Beyond raw performance, LLaMA~3.2 Vision offers the added advantage of interpretability through physics-grounded textual explanations, enabling the model to articulate reasoning tied to event topologies such as muon tracks, electromagnetic showers, or the absence of visible charged-lepton signatures. This capacity for explainable predictions is particularly valuable in scientific workflows, where transparent decision-making facilitates trust, debugging, and integration with expert knowledge. We emphasize that these explanations are post hoc in nature and are not claimed to reflect the model’s internal causal reasoning; nonetheless, they provide an accessible interface for qualitative inspection and error analysis that is not available in typical CNN or ViT-based approaches.

Our ablation study further shows that removing explicit physics definitions from the system prompt does not significantly degrade classification performance, indicating that the fine-tuned model relies primarily on learned visual–semantic representations rather than prompt-level domain descriptions. This suggests that the vision–language model internalizes task-relevant features during fine-tuning, while still leveraging its language generation capabilities to produce plausible physics-aware explanations.

In contrast, a few-shot in-context evaluation using the frozen pre-trained VLM fails to yield meaningful class separation, demonstrating that prompt-based adaptation alone is insufficient for mapping sparse detector images to abstract physical interaction categories. This result underscores the necessity of parameter adaptation, even when using large pretrained vision–language models, for specialized scientific imaging tasks.

Taken together, these results suggest a natural hierarchy of model choices for neutrino event classification. CNNs retain an important role in scenarios requiring real-time inference or operation under strict resource constraints, such as on-detector edge computing or rapid online filtering. Vision transformers provide an effective intermediate solution, combining strong classification performance and robustness with moderate computational requirements for large-scale offline processing. In contrast, vision–language models are especially well-suited for offline analyses and detailed event studies in neutrino physics, where interpretability, adaptability, and robustness to detector variations are as critical as raw accuracy.

Looking ahead, promising research directions include compressing large transformer models through quantization and pruning, distilling vision-language models into compact architectures that retain interpretability, and developing domain-specific foundation models trained on diverse neutrino event topologies. Such efforts could enable reusable physics foundation models that generalize across detector configurations and experiments with minimal fine-tuning. In addition, a more detailed and systematic analysis of post hoc explanations is essential to assess their physical fidelity, consistency, and practical usefulness for domain experts. Future work may also explore fine-tuning strategies that explicitly incorporate explanatory objectives or supervision, with the goal of improving the alignment between model predictions, generated explanations, and established physical reasoning. These developments would help bridge the gap between the accuracy and explainability of large-scale models and the efficiency of lightweight architectures, bringing the benefits of transformer-based and multimodal approaches to a wider range of deployment environments in experimental physics.

\begin{table}[!h]
    \caption{Event classification aggregated metrics.}
    \label{metrics_table}
    \centering
    \begin{tabular}{lccc}
    \toprule
        Metric & LLaMA 3.2 Vision & ViT-h/14 & CNN\\
    \midrule
         Accuracy & 0.87 & 0.86 & 0.80 \\
         Precision & 0.87 & 0.86 & 0.80 \\
         Recall & 0.87 & 0.85 & 0.79 \\
         AUC-ROC & 0.96 & 0.96 & 0.94 \\
     \midrule
     \# of Trainable Parameters &29.5M (QLoRA)& 632M & 21.7M\\
     Training Regime & PEFT, 1 epoch & Full, 10 epochs & Full, 300 epochs\\
     \midrule
         Inference Memory Usage (GB) & 12.91 &2.56& 2.44\\
         Time per Sample (ms) & 3412 & 299.1 & 23.90\\
    \bottomrule
    \end{tabular}\\
\end{table}

\begin{table}[!h]
    \caption{Event classification aggregated metrics for generalization testing.}
    \label{metrics_table_generalization}
    \centering
    \begin{tabular}{lccc}
    \toprule
        Metric & LLaMA 3.2 Vision & ViT-h/14& CNN \\
    \midrule
         Accuracy & 0.85 & 0.85 & 0.43 \\
         Precision & 0.85 & 0.85& 0.4 \\
         Recall & 0.85 & 0.85&0.41 \\
         AUC-ROC & 0.95 & 0.96&0.70 \\
    \bottomrule
    \end{tabular}\\
\end{table}

\section*{Funding Statement}
This work was supported by the U.S. Department of Energy under Award Number DE-SC0009920 awarded to J.B.

\section*{Author Contributions}
D.S. developed the VLM and ViT pipeline, fine-tuned the VLM and ViT models, and ran all VLM and ViT experiments. K.Y. implemented and trained the CNN baselines. A.Y. prepared and curated the datasets. J.B. and P.B. conceived the project and provided guidance and funding. All authors contributed to manuscript drafting, review, and editing.

\section*{Competing interests}
The Authors declare no competing interests.

\section*{Data availability}
The detector pixel map data that support the findings of this study are available from the corresponding author, J.B., upon request.

\section*{Code Availability}
The code used in this study is made available at: 
\url{https://github.com/dikshantsagar/Neutrino-LLaMa}.

\end{document}